\UseRawInputEncoding
\documentclass[a4paper,twoside]{article}
\usepackage{enumitem}
\usepackage{float}
\usepackage{amsfonts}
\usepackage{amssymb} 
\usepackage{amsmath}
\usepackage{amsthm}
\usepackage{hyperref}
\usepackage[nameinlink,capitalize]{cleveref}
\usepackage{svg}
\usepackage[dvipsnames]{xcolor}

\usepackage{graphicx}
\usepackage{subcaption}
\usepackage{algorithm}
\usepackage{algpseudocode}
\usepackage[algo2e]{algorithm2e}
\usepackage{lipsum}
\usepackage{makecell}
\usepackage{lastpage}
\usepackage{booktabs}

\usepackage{multirow}

\usepackage{apalike}
\usepackage{SCITEPRESS}

\begin{document}

\title{MEDALLION: Manifold Embedding Distillation and Attributions for Low-dimensional Latent InterpretatiON}

\author{
\authorname{
Vicky Jung\sup{2},
Claire M. He\sup{1},
Irene Chang\sup{1} and
Genevera I. Allen\sup{1}
}
\affiliation{\sup{1}Department of Statistics, Columbia University, New York, USA}
\affiliation{\sup{2}University of Florida, Gainsville, USA}
}


\abstract{
Dimensionality reduction is widely used to visualize high-dimensional data and guide scientific discovery, yet interpreting which original features contribute to low-dimensional representations remains challenging, particularly for black-box methods such as t-SNE and UMAP. We introduce \textbf{MEDALLION} (Manifold Embedding Distillation and Attributions for Low-dimensional Latent InterpretatiON), a novel framework for interpreting low-dimensional embeddings through knowledge distillation and neural feature attribution. MEDALLION distills any given embedding with a neural encoder which we explain via a broad range of existing neural feature attribution methods. Through experiments on MNIST and applications in single-cell transcriptomics, we show MEDALLION provides informative feature attributions that cast additional insights into the underlying embedding and compare its performance with existing interpretability approaches. 
}

\onecolumn
\maketitle
\normalsize
\setcounter{footnote}{0}
\vfill

\section{INTRODUCTION}
Dimensionality reduction (DR) is widely used in the applied sciences to summarize and visualize complex data: high-dimensional observations in $\mathbb{R}^p$ are mapped onto a lower-dimensional embedding space $\mathbb{R}^d$, with $d \ll p$ (typically with $d=2$ or $3$ for visualization).  By preserving selected properties of the original data, these embeddings reveal structures such as clusters and trajectories that motivate scientific hypotheses and downstream analyses across domains ranging from market analysis to biomedicine. Numerous methods have been developed to construct such representations, from Principal Component Analysis (PCA, \cite{hotelling_analysis_1933}) and Locally Linear Embedding (LLE, \cite{roweis_nonlinear_2000}), to modern black-box methods such as t-SNE \cite{maaten_visualizing_2008} and UMAP \cite{mcinnes_umap_2020}. However, these DR methods produce embeddings that distort the original data \cite{chari_specious_2023,wang_what_2023} and are sensitive to preprocessing, initialization, hyperparameters, and feature selection \cite{lin_calibrating_2024,wang_understanding_nodate,huang_towards_2022,nguyen_ten_2019}, raising concerns about reliability and reproducibility \cite{bodt_low-dimensional_2025,sun_accuracy_2019,jeon_stop_2026}, especially for the downstream conclusions. Understanding which original features contribute to the resulting embeddings is therefore important for interpreting and assessing the scientific conclusions drawn from them.
\medskip

\noindent Feature attribution methods in Interpretable Machine Learning (IML) \cite{allen_interpretable_2023,murdoch_definitions_2019} seek to explain embeddings by identifying and scoring the contributions of original features. Several model-agnostic approaches adapt local surrogate methods from supervised learning to explain embedding coordinates, including BIR, LXDR, and LIMEADE \cite{bibal_explaining_nodate,bardos_local_2022,zikry_limeade_2025}. Related approaches, such as SLISEMAP \cite{bjorklund_slisemap_2023}, jointly learns embeddings and locally interpretable models. However, these explanations depend on the fidelity of the local nonlinear approximation and can be highly sensitive to neighborhood selection and surrogate hyperparameters. Other approaches have sought to explain popular DR algorithms such as t-SNE and UMAP directly through gradients \cite{corbugy_gradient-based_2024}, genetic programming \cite{lensen_genetic_2021}, or visualization diagnostics \cite{ma_visual_2021,zeng_ihelp_2023}.  While these approaches can provide more direct insights into particular DR algorithms, they require reformulations of the original optimization problem or computationally intensive searches.

\medskip

\noindent  Originally developed for compressing complex \emph{teacher} models to simpler and cheaper \emph{student} models \cite{hinton_distilling_2015,bucilua_model_2006}, knowledge distillation (KD) has since been used for interpretability, e.g distilling black-box predictions to interpretable models such as decision trees \cite{frosst_distilling_2017,zhou_approximation_2018}. Subsequent work has examined student-teacher explanations \cite{tan_distill-and-compare_2018,bastani_interpretability_2018,han_impact_2023,moreira_diconstruct_2024} and explanation-aware KD that preserves feature attributions alongside predictions \cite{alharbi_learning_2021,chowdhury_improving_2023,parchami-araghi_good_2024,sun_explainability-based_2025}. Although KD has also been explored for explaining unlabeled neural network representations \cite{zhang_unsupervised_2018}. its use for post-hoc feature attribution in unsupervised learning remains largely underexplored. \medskip

\paragraph{Contributions}
We introduce \textbf{MEDALLION} (Manifold Embedding Distillation and Attributions for Low-dimensional Latent InterpretatiON), a general framework for interpreting low-dimensional embeddings via knowledge distillation and neural feature attribution. We use a distilled neural encoder that can be explained by a broad class of neural-network-based feature attribution methods. We further consider MEDAL \cite{chang_medal_2026} as a particular instantiation of MEDALLION, in which the encoder is jointly trained with a decoder under an additional reconstruction objective. Through the MNIST dataset and two real-data applications in transcriptomics, we evaluate the ability of MEDALLION to identify informative features underlying low-dimensional representations and compare its performance with existing approaches for DR interpretability.
\section{EXPLAINING EMBEDDINGS THROUGH ENCODER DISTILLATION}

Our objective is to identify which original features contribute to the low-dimensional embedding produced by a given DR method. To this end, we propose to distill the observed embedding into a neural encoder, allowing us to leverage existing neural-network-based feature attribution methods for interpretation.

\paragraph{Distilling the observed embedding.}
Let $X = \{x_i\}_{i=1}^n \subset \mathbb{R}^p$ denote the training data, and let $f : \mathbb{R}^{n \times p} \to \mathbb{R}^{n \times d}$, with $d \ll p$, denote a DR procedure such as PCA, t-SNE, or UMAP, producing a low-dimensional embedding $Z=f(X)$. For each observation $x_i$, the DR procedure produces a corresponding embedding coordinate $z_i = [f(X)]_i \in \mathbb{R}^d$. We consider a neural encoder $\hat f_\theta : \mathbb{R}^p \to \mathbb{R}^d$, parametrized by $\theta$, trained to reproduce the observed embedding by minimizing the empirical distillation loss
\begin{equation}
\min_\theta \mathcal{L}_d(\theta)
=
\frac{1}{n}\sum_{i=1}^n
\|\hat f_\theta(x_i)-z_i\|_2^2.
\label{eq:dist}
\end{equation}
With a sufficiently expressive encoder, the empirical distillation error can be made arbitrarily small given the appropriate architecture and training, providing an explicit map that interpolates the observed embedding perfectly on the training data. Moreover, the learned encoder defines a mapping $x_{\mathrm{new}} \mapsto \hat f_\theta(x_{\mathrm{new}})\in\mathbb{R}^d$ for previously unseen observations, extending the observed representation without refitting the original DR procedure and offering the ability to give feature attributions to explain both in-sample observations and held-out observations.  

\paragraph{Feature attributions via a distilled encoder.}
A key advantage of the distilled encoder is that its neural architecture accommodates a broad class of feature attribution methods, including Integrated Gradients (IG, \cite{sundararajan_axiomatic_2017}), Layer-wise Relevance Propagation (LRP, \cite{villa_layer-wise_2016}), DeepLIFT \cite{shrikumar_learning_2019}, and SmoothGrad \cite{smilkov_smoothgrad_2017}, among others. In particular, for each embedding coordinate $l \in\{1,\ldots,d\}$, where $d=2$ for uses of DR in data visualization, we apply a feature attribution $\mathcal{I}$ method to the scalar-valued mapping $x \mapsto \{\hat f_{\theta}(x)\}_{l} \in \mathbb{R}$ to obtain
\[
\mathcal{I}^{(l)}(x) = \bigl(\mathcal{I}_{1}^{(l)}(x),\ldots, \mathcal{I}_{p}^{(l)}(x)\bigr) \in \mathbb{R}^p,
\]
where $\mathcal{I}_{j}^{(l)}(x)$ characterizes the contribution of the $j$-th original feature to the $l$-th embedding coordinate according to the selected attribution method. For example, given a baseline $x_0$, Integrated Gradients assigns
\begin{align}
\mathcal{I}_{j, \operatorname{IG}}^{(l)}(x;x_0) &= \nonumber \\
(x_j-x_{0j}) &\int_0^1 \frac{\partial \bigl(\hat f_{\theta}
\bigl(x_0+t(x-x_0)\bigr)\bigr)_{l}}{\partial x_j}
\,dt.
\end{align}

\noindent Our framework therefore provides coordinate-wise feature attributions that can be summarized to feature scores  $\Delta^{(j)}(x)= ||\mathcal{I}_j(x)||_1$, for $x$ both in-sample and previously unseen, allowing us to investigate which original features contribute to the learned low-dimensional representation establishing the embedding and which original features might explain how we embed new data onto this lower dimensional representation. 

\paragraph{Application via MEDAL \cite{chang_medal_2026}.}
MEDAL provides a particular instantiation of our framework in which the distilled encoder $\hat f_\theta : \mathbb{R}^p \to \mathbb{R}^d$ is jointly trained with a decoder $\hat g_\phi : \mathbb{R}^d \to \mathbb{R}^p$ to reconstruct the original observations. In addition to matching the teacher embedding exactly like in our proposal, MEDAL also incorporates a reconstruction objective to encourage the encoder to retain information from the original feature space. Specifically, MEDAL minimizes  Specifically, MEDAL minimizes
\begin{align}
\min_{\theta,\phi} \mathcal{L}(\theta,\phi)
= \min_{\theta, \phi} \lambda_d &\mathcal{L}_d(\theta) \; + \nonumber \\ &\mathcal{L}_r(\theta, \phi),
\quad \lambda_d \geq 0.
\end{align}
with distillation loss $\mathcal{L}_d(\theta) = \frac{1}{n}\sum_{i=1}^n
\|\hat f_\theta(x_i)-z_i\|_2^2$ same as \autoref{eq:dist} and reconstruction loss
$\mathcal{L}_r(\theta, \phi) = \frac{1}{n}\sum_{i=1}^n
\|x_i-\hat g_\phi(\hat f_\theta(x_i))\|_2^2$. MEDAL uses a sufficiently large $\lambda_d$ prioritizes driving the distillation loss toward zero while minimizing reconstruction error. Conceptually, this corresponds to minimizing reconstruction error subject to reproducing the teacher embedding exactly. Within our MEDALLION framework, we apply feature attribution methods to the distilled encoder rather than to the full encoder-decoder mapping or its reconstruction error, as the latter would characterize features contributing to the reconstruction of the original observations rather than specifically to their low-dimensional representation. In particular, reconstruction-based attributions may reflect both the information preserved in the embedding and the decoder’s ability to recover the original features, thereby conflating distinct sources of variation. Nevertheless, the decoder provides complementary reconstruction diagnostics, allowing us to investigate how feature attributions differ between well- and poorly reconstructed observations, including previously unseen samples.

\section{RESULTS}
\subsection{Benchmark MNIST-64 Dataset}
\begin{figure*}[t]
    \centering
    \includegraphics[width=\linewidth]{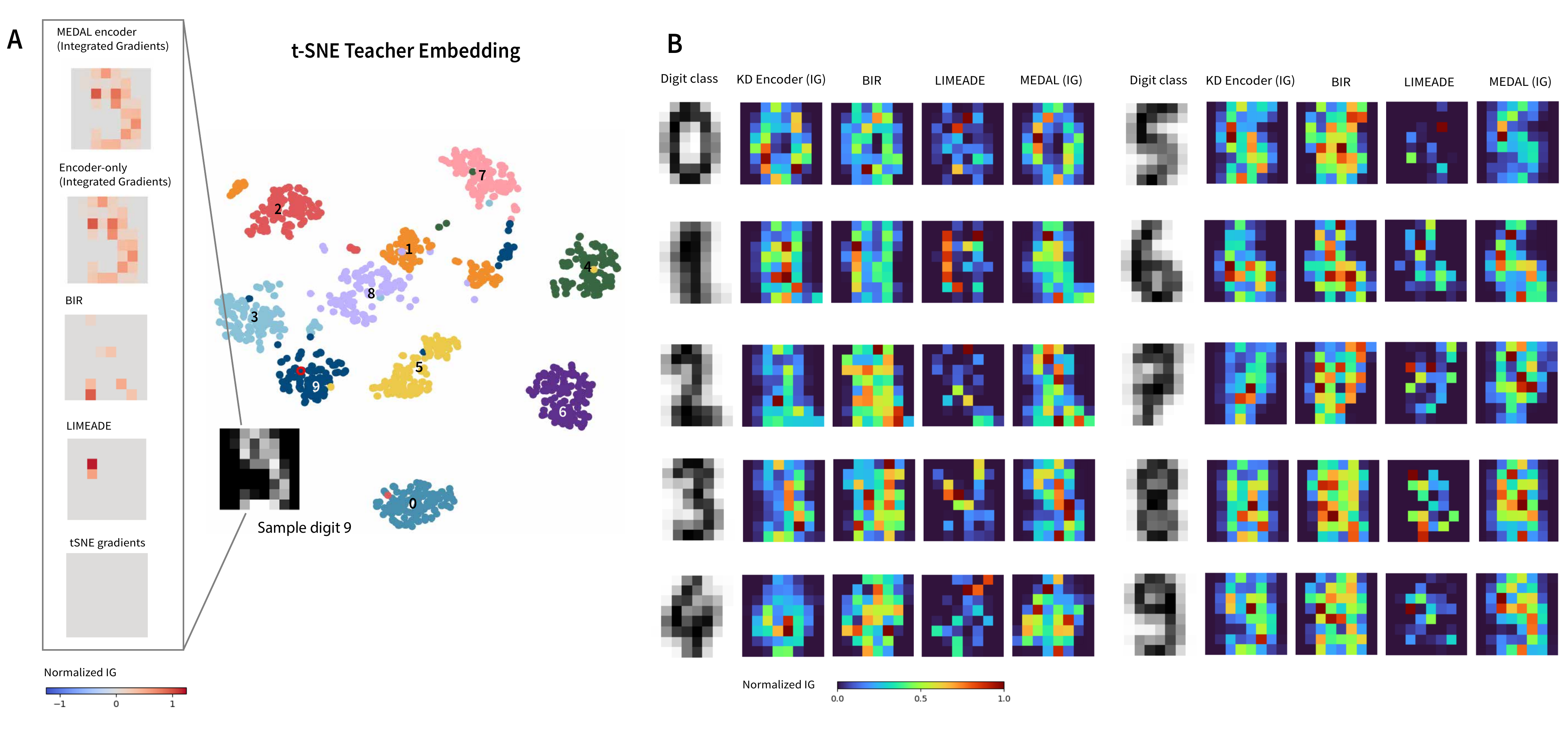}
        




    \caption{\textbf{MNIST-64's t-SNE embedding's feature attributions. }\textbf{A.} Pixel-level feature importance for a sample of digit 9. \textbf{B.} Per-class averaged feature attributions across different methods: Encoder-only IG, BIR, LIMEADE and MEDAL IG obtained on training data used for the embedding.}
    \label{fig:mnist_combined}
\end{figure*}

\begin{figure*}
    \centering
    \includegraphics[width=0.9\linewidth]{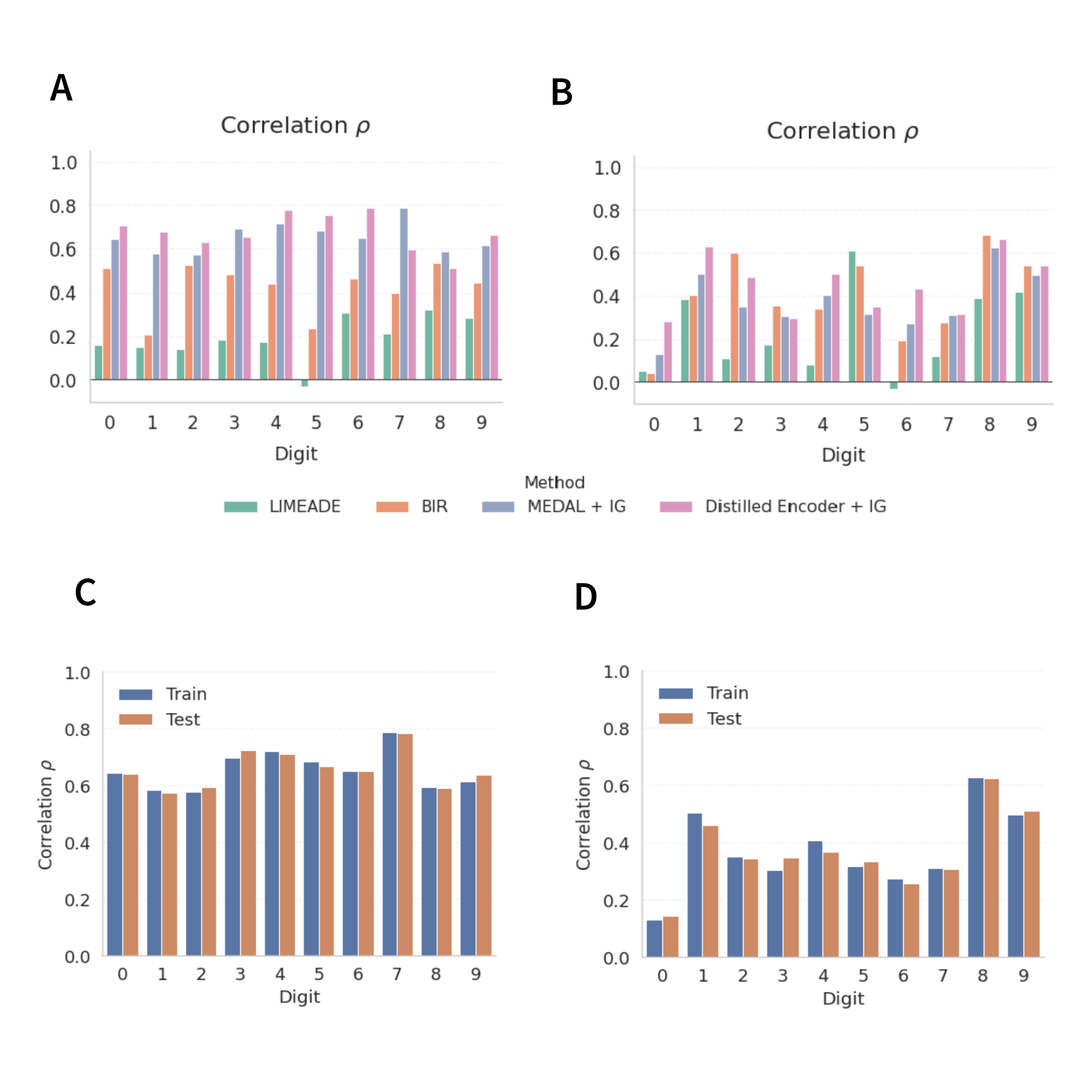}
    \caption{\textbf{Per digit-class feature attribution validation.} \textbf{A}. Correlation of DR feature attributions with supervised MLP with IG. \textbf{B}. Correlation of DR feature attributions with supervised Random Forest MDI. \textbf{C}. Correlation of MEDAL with IG with supervised MLP with IG for training samples and test samples. \textbf{D}. Correlation of MEDAL with IG with supervised Random Forest MDI for training samples and test samples.}
    \label{fig:mnist-add}
\end{figure*}

\paragraph{Digit-class feature importance}
We first assess the capabilities of our framework to recover meaningful feature attributions in a classic IML example with the simple MNIST 10 digits 64-pixel dataset. We choose as teacher embedding a t-SNE teacher tuned using MEDAL's hyperparameter selection scheme ie with a perplexity value that minimizes held-out reconstruction error on the distilled embedding with a train/test sample split of 80/20\%. We compare the results of different feature attribution methods in the literature, namely BIR \cite{bibal_explaining_nodate}, LIMEADE \cite{zikry_limeade_2025} and t-SNE gradients \cite{corbugy_gradient-based_2024}. We find our distilled encoder approach (with encoder only or MEDAL) yields the best qualitative feature attributions across number classes (Fig.~\ref{fig:mnist_combined}\textbf{A}) and in individual samples. Additionally, in order to validate quantitatively that the feature attributions produced with our methods outperform the existing methods, we evaluate the agreement between the unsupervised feature attributions with two supervised experiments computing the following feature attributions: 1) the Mean Decrease in Impurity (MDI) of a supervised classifier (Random Forest Classifier tuned at a 96.88\% test accuracy), and 2) the Integrated Gradients for an MLP supervised classifier (tuned at a 98.96\% test accuracy), on the same data, using the digit class as label. The per-sample and per-class correlations are reported in the additional materials (Table~\ref{tab:mnist-global} and Fig.~\ref{fig:mnist-add}), where we find that our approaches with a distilled encoder or via MEDAL outperform existing methods on most digit classes and on average at sample level. Moreover, our approach yields stable per-digit class correlation across train and test splits and can be reliably used for test samples. 

\begin{figure*}[t]
    \centering
   \includegraphics[width=\linewidth]{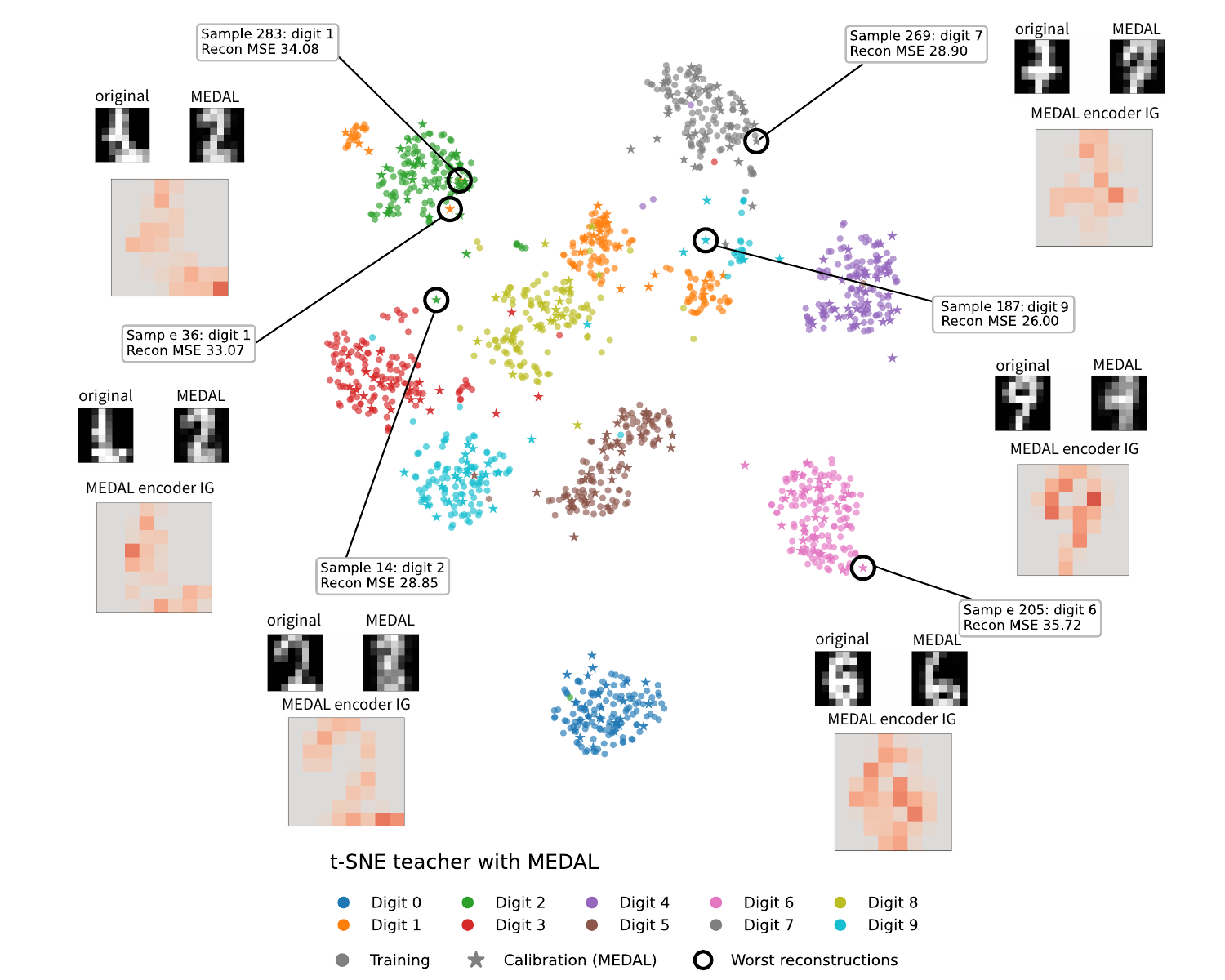}
    \caption{\textbf{MNIST-64's poorly reconstructed test samples.} We highlight the 6 worst reconstructed test samples on the embedding and their MEDAL reconstruction and respective feature attributions (IG).}
    \label{fig:MNIST_poorly}
\end{figure*}

\paragraph{Feature importance of poorly reconstructed samples.}
Coupled with MEDAL's reconstruction mapping to detect embedding distortions, our framework enables feature-level investigation of poorly reconstructed observations. By aggregating local feature importance scores over these observations, we identify features that drive their low-dimensional representations and may contribute to reconstruction discrepancies. This provides a means to investigate which aspects of the original data are associated with distortions in the learned embedding. We highlight in Fig.~\ref{fig:MNIST_poorly} the 6 worst reconstructed samples in the test dataset with their feature attributions. For instance, samples 283 and 36 of a digit 1 are reconstructed by MEDAL as a 2 and embedded close to the samples corresponding to digit class 2, their corresponding feature attribution maps show that the features driving this reconstruction are pixels associated with digit class 2 but not digit class 1 (e.g. pixel at column 7, row 8). Beyond samples who are reconstructed incorrectly by MEDAL, our method also serves to identify features responsible for uncertainty in the reconstruction of the samples: samples 269 and 187 are correctly reconstructed, however the digits depart from the canonical digit class representation (see Fig.~\ref{fig:mnist_combined}\textbf{B}) and the attribution highlights which features of the sample contributes to a misleading reconstruction. 

\section{APPLICATIONS TO TRANSCRIPTOMICS DATASETS}

In this section, we demonstrate two applications of MEDALLION to transcriptomics datasets. Single-cell transcriptomics is often high-dimensional, making low-dimensional representations common to explore and investigate the data. These two applications in the immune cells of a cancer study \cite{jerby-arnon_cancer_2018} and a mouse brain study \cite{tasic_shared_2018} reveal that major gene signatures are preserved in low dimensional embeddings.

\subsection{Immune Dataset In A Cancer Study}
\begin{figure*}[t]
    \centering
    \includegraphics[width=\linewidth]{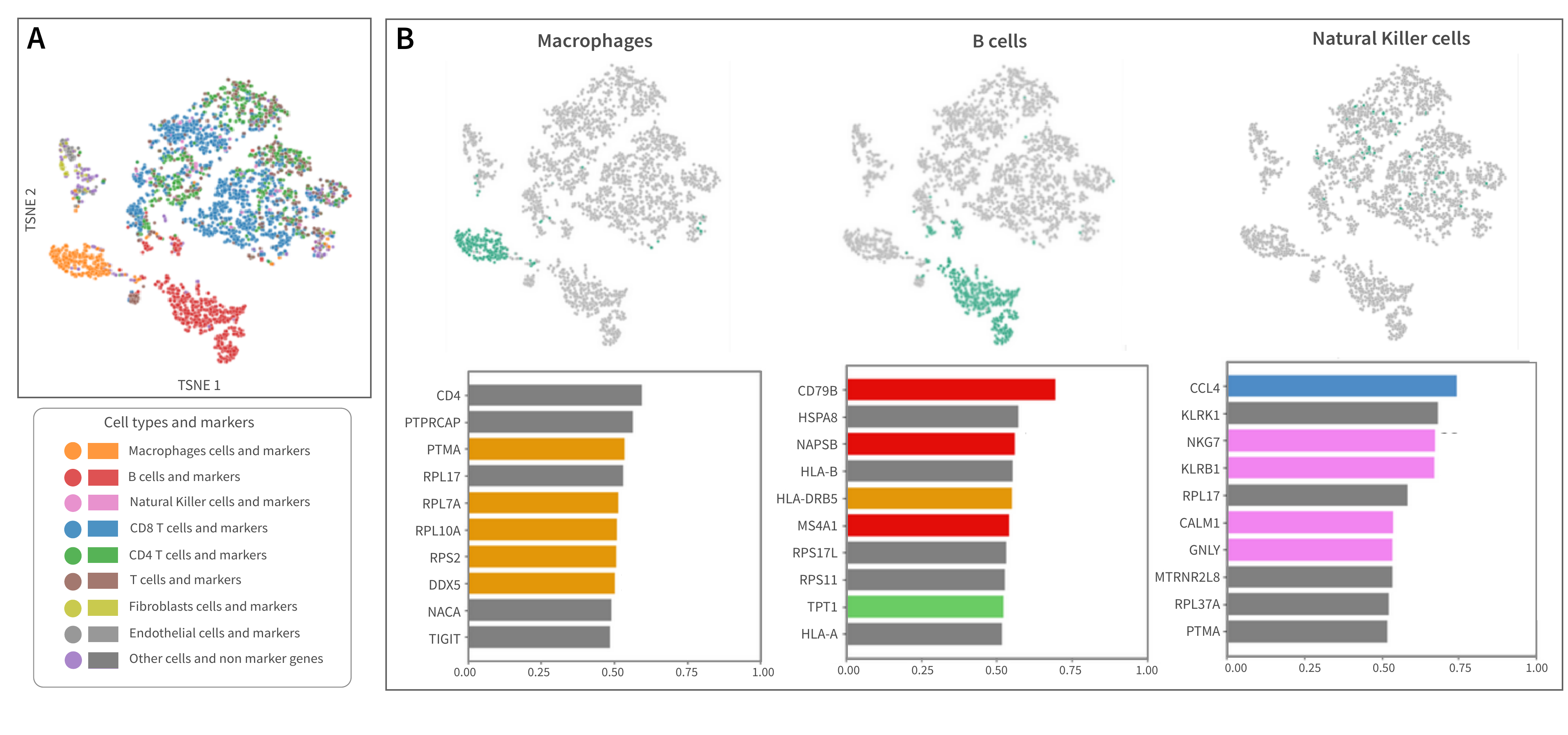}
    \caption{\textbf{T-SNE embeddings and feature attributions} for the immune cell dataset from a cancer study \cite{jerby-arnon_cancer_2018}. \textbf{A.} T-SNE embedding colored with the original paper's cell type labels. \textbf{B.} MEDALLION's top 10 feature attributions for Macrophages, B cells and Natural Killer cells.}
    \label{fig:cancer_data}
\end{figure*}

Our first dataset of immune cells collected in a cancer study \cite{jerby-arnon_cancer_2018} is extracted from a dataset containing 3292 tumor cell profiles of three cell lines of melanoma patients. This data was separated into two compartments of malignant or stromal cells (CD45- cells) and immune cells (CD45+ cells). We focused on the low-dimensional representation of the immune cells which after preprocessing included 6912 genes. Following the original paper, the teacher for this low-dimensional embedding is a t-SNE embedding computed at default hyperparameters (with perplexity value 30) for which we distilled MEDAL as shown in Fig.~\ref{fig:cancer_data}\textbf{A} with the cell type labels from the study. 

\paragraph{MEDALLION retains known marker genes for cell types}
Using Integrated Gradients on the distilled MEDAL encoder, we assess the feature attributions of the learned embedding. In Fig.~\ref{fig:cancer_data}\textbf{B}, MEDALLION identifies known gene markers in the top 10 highest feature attributions for particular cell types: these genes contribute the most to regions of the embedding associated with certain cell types. B cells in particular delineate a particular region embeddings for which MEDALLION's attributions highlight \textit{CD79B} or \textit{MS4A1}, major B cell gene markers. Even in the absence of a distinct region for a particular cell type like Natural Killer cells in the embedding, MEDALLION's top attributions of \textit{NKG7} and \textit{GNLY} which are canonical markers of the Natural Killer cell type reveal that the embedding retains this gene information. 

\subsection{A Mouse Cortical Transcriptomics Dataset}
\begin{figure*}[t]
    \centering
    \includegraphics[width=\linewidth]{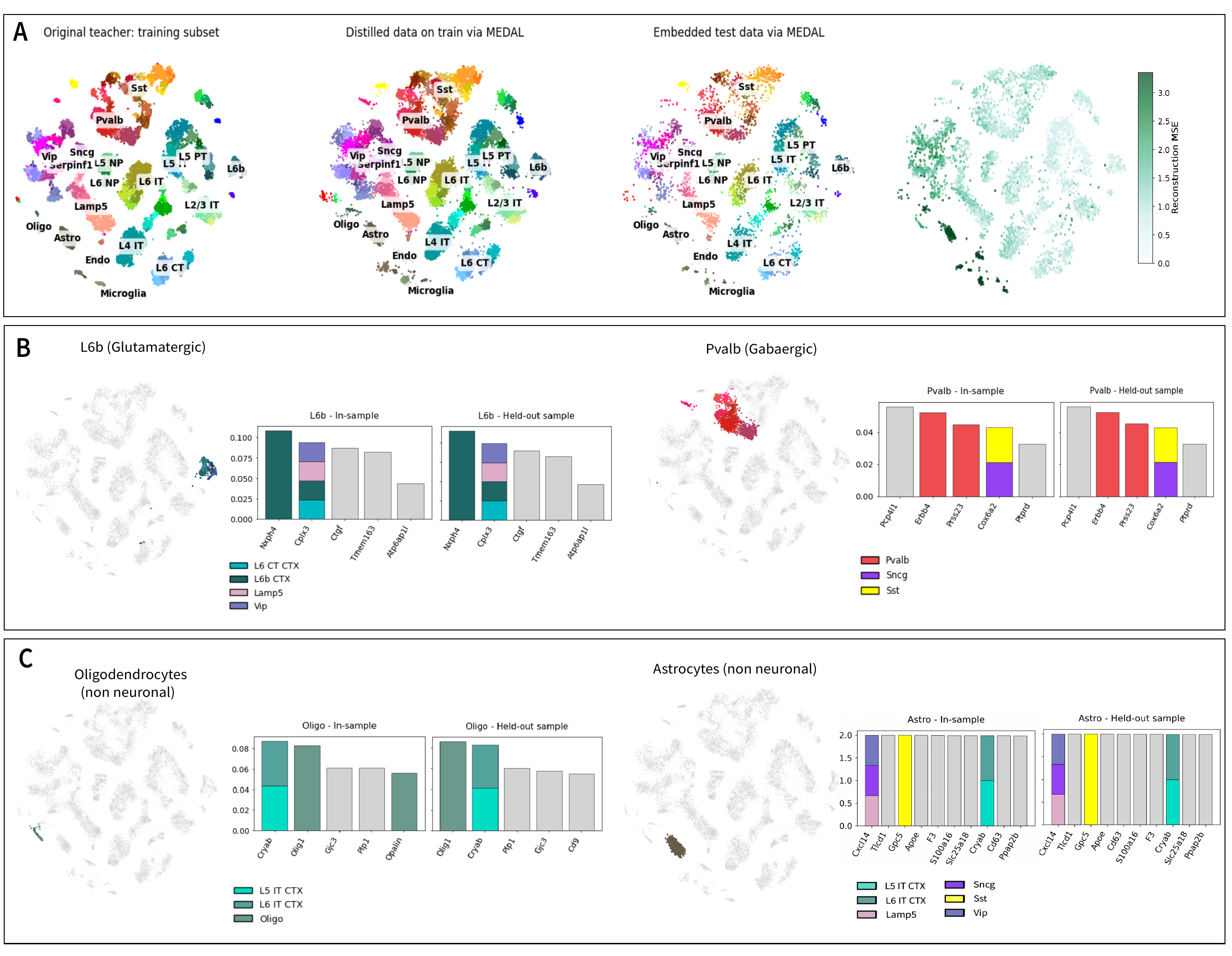}
    \caption{\textbf{T-SNE embeddings and top-5 feature attributions} for the mouse neo-cortex data \cite{tasic_shared_2018}. \textbf{A}. From left to right: Original teacher embedding using t-SNE, MEDAL distilled embeddings on training split, MEDAL's embedded test data, Test sample reconstruction error map. \textbf{B}. L6b and Pvalb neuron subclasses and \textbf{C}. Oligodendrocytes and Astrocytes (top 10) subclasses feature attributions, known gene markers are colored based on \cite{yao_high-resolution_2023}.}
    \label{fig:tasic_scatter_MEDAL}
\end{figure*}

Our second dataset from Tasic et al. \cite{tasic_shared_2018} contains 23,800 cells from the adult mouse brain with a reported 133 cell types hierarchically organized (see Fig.~\ref{fig:tasic_scatter_MEDAL}\textbf{A}). We reproduced the exact tSNE obtained by \cite{bodt_low-dimensional_2025} as teacher for MEDAL explained with Integrated Gradients after perfect distillation of the encoder. As established by \cite{bodt_low-dimensional_2025}, the t-SNE teacher embedding suggests some well separated families (as labelled) but do not reveal the hierarchical structure inferred by the original study via their clustering pipeline. 

\paragraph{Cell type structure via feature attributions} We grouped the 133 cell types of the original dataset into 18 known meta-cell types for which we used gene marker annotations from the Allen Brain Atlas \cite{yao_high-resolution_2023} which correspond to the cortex subclasses. Known subclasses of both Glutamatergic neurons L6b and Gabaergic Pvalb have top feature attributions consistent with their annotations (Fig.~\ref{fig:tasic_scatter_MEDAL}\textbf{B}) and as evidenced in \cite{sanzmorello_complex_2020}, the Sncg/Sst attributed \textit{Cox6a2} gene marker also shows overlap of expression with the Pvalb subclass.  Additionally, MEDAL's distortion map reveals that the distorted samples belong to major non neuronal cell types (Oligodendrocytes, Astrocytes, Endothelial cells and Microglia) that are present in the mouse neo-cortex, while samples in L2/3, L4, L5, L6 (subclass of glutamatergic neurons) are some of the better reconstructed samples on the embeddings. In Fig.~\ref{fig:tasic_scatter_MEDAL}\textbf{C}, among the poorly reconstructed non-neuronal cells, Oligodendrocytes are some of the better represented cells in the embeddings, and MEDALLION identifies known marker genes in the top attributions. In contrast, Astrocytes exhibit higher reconstruction errors, and their top ten attributed features - here obtained via rank-percentile aggregation due to low normalized scores, do not include consistent cell-type markers, suggesting that caution is warranted when deriving annotations for these samples.

\section*{DISCUSSION}
We introduced MEDALLION, a general framework for interpreting \textit{any low-dimensional embeddings} through knowledge distillation and neural-network-based feature attribution. By distilling a given embedding into a neural encoder, MEDALLION makes existing neural feature attribution methods applicable to embeddings produced by diverse DR algorithms. The learned encoder further extends feature attribution to previously unseen observations, allowing the interpretation of both in-sample and held-out data within a common embedding space.

\noindent Our experiments demonstrate the potential of this approach to identify informative features underlying low-dimensional representations. On MNIST, integrated gradients obtained through the distilled encoder produces interpretable attributions of digit-specific pixels to each digit-class. Both the distilled encoder and MEDAL encoder show stronger agreement with supervised feature attributions than other existing methods. In transcriptomics datasets, MEDALLION provides additional insights for discovery of cell-type and cell-type markers derived from low-dimensional embeddings. Our method provide complementary evidence of discovery for certain cell types and help identify unreliable findings in poorly reconstructed samples.

\noindent Importantly, as a surrogate model, MEDALLION interprets the \textit{observed embedding} through a distilled encoder, rather than directly explaining the original DR algorithm. Accurate reproduction of embedding coordinates does not necessarily mean that the student encoder uses the same input features in the same way as the original DR method, a distinction between predictive and explanatory fidelity already recognized in the knowledge distillation literature \cite{wu_what_2024}. 
In DR, however, assessing explanatory fidelity is challenging because there are generally no ground-truth feature attributions against which to evaluate the surrogate's explanations.
MEDALLION provides a principled and flexible approach to estimating feature attributions from the observed embedding, without requiring direct explanations of the original DR procedure.  



\noindent Beyond cell-type discovery and marker annotation validation in single-cell data analysis, MEDALLION can also investigate more complex biological structures including continuous cell-state transitions or differentiation trajectories in development (e.g. highlighting important feature attributions for PHATE embeddings \cite{moon_visualizing_2019}). Moreover, data integration of multi-modal transcriptomics datasets often hinges on embedding parts or all of the datasets into a common low-dimensional representation. In these settings, feature attributions provided by MEDALLION and the ability to map new samples could yield insights into the shared representations and on the retained biological signals. 

\noindent More broadly, MEDALLION complements the MEDAL framework that extends dimension reduction as a visual exploratory tool towards a quantitative part of the scientific workflow by providing feature-level explanations. While it might explain features contributing to patterns compressed in visual embeddings, MEDALLION should not be treated as a means to infer causal relationship or replace careful differential expression analysis. Rather, MEDALLION is a step towards a more transparent and trustworthy workflow of single-cell data analysis.


\section*{ADDITIONAL MATERIALS}

  \paragraph{Implementation details} MEDALLION encoder and MEDAL architectures were chosen following the MEDAL default architecture search described in \cite{chang_medal_2026}. 
  
\paragraph{Comparison with other methods} We compared our methods against three of the feature attributions methods for DR, namely BIR, LIMEADE and t-SNE Gradients for our embeddings obtained using t-SNE. We note that BIR had \href{https://github.com/vu-minh/mlteam-lime-for-tsne/blob/master/}{original code available in Python}, LIMEADE and t-SNE gradients were re-implemented following the algorithms descriptions in their respective papers. In our experiments, t-SNE gradients were especially expensive to compute (14h) and returned near zero explanations everywhere - therefore were not included in the downstream analysis. For LIMEADE, choice of $\gamma \in [0.1, 0.01, 0.001, 0.0001]$ was made to obtain the best digit-class feature attributions.

\paragraph{Data processing} The immune cell dataset \cite{jerby-arnon_cancer_2018} was processed according to the original paper's processing steps to reproduce the original embedding. The mouse neo-cortical dataset \cite{tasic_shared_2018}
 was processed as outlined by \cite{bodt_low-dimensional_2025}.

\begin{table*}
    \centering
    \caption{Pearson and Spearman correlations between supervised and 
    unsupervised per-sample feature attributions on MNIST.}
    \label{tab:mnist-global}
    \vspace{0.2cm}
    \begin{tabular}{llcccc}
        \toprule
        & & \multicolumn{4}{c}{Unsupervised feature attribution} \\
        \cmidrule(lr){3-6}
        Metric & \makecell{Supervised \\ reference }
        & MEDAL + IG & KD Encoder + IG & BIR & LIMEADE \\
        \midrule
        \multirow{2}{*}{Pearson $\rho$}
            & MLP + IG  & 0.342 (0.002) &\textbf{ 0.377} (0.005) & -0.010 (0.004) & -0.016 (0.006) \\
            & RF + MDI  & 0.291 (0.004) & \textbf{0.355} (0.004) &  0.114 (0.004) & 0.04 (0.005) \\
        \addlinespace[0.5em]
        \multirow{2}{*}{Spearman $\rho$}
            & MLP + IG  & 0.307 (0.004) & \textbf{0.314} (0.004) & 0.028 (0.005) & 0.008 (0.007) \\
            & RF + MDI  & 0.483 (0.003) & 0.509 \textbf{(0.003)} & 0.23 (0.003) & 0.113 (0.004)\\
        \bottomrule
    \end{tabular}
\end{table*}

\bibliographystyle{apalike}
\bibliography{references}

\end{document}